\documentclass{article}

\usepackage{microtype}
\usepackage{graphicx}
\usepackage{subcaption}
\usepackage{booktabs} 

\usepackage{hyperref}

\usepackage[preprint]{icml2026}

\usepackage{amsmath}
\usepackage{amssymb}
\usepackage{mathtools}
\usepackage{amsthm}
\usepackage{multirow}
\usepackage[table]{xcolor}
\usepackage{color}
\definecolor{mygreen}{RGB}{146, 199, 113} 
\usepackage{graphicx}
\usepackage{adjustbox}
\usepackage{array}
\usepackage{booktabs}
\usepackage{dsfont}
\usepackage{makecell}

\usepackage[capitalize,noabbrev]{cleveref}

\theoremstyle{plain}

\theoremstyle{definition}

\theoremstyle{remark}

\usepackage[textsize=tiny]{todonotes}

\icmltitlerunning{Self-Supervised Weight Templates for Scalable Vision Model Initialization}

\begin{document}

\twocolumn[
  \icmltitle{Self-Supervised Weight Templates for Scalable Vision Model Initialization}



  \icmlsetsymbol{equal}{*}

  \begin{icmlauthorlist}
    \icmlauthor{Yucheng Xie}{add1,add2}
    \icmlauthor{Fu Feng}{add1,add2}
    \icmlauthor{Ruixiao Shi}{add1,add2}
    \icmlauthor{Jing Wang}{add1,add2}
    \icmlauthor{Yong Rui}{add1,add2}
    \icmlauthor{Xin Geng}{add1,add2}
  \end{icmlauthorlist}

  \icmlaffiliation{add1}{School of Computer Science and Engineering, Southeast University, Nanjing, China}
  \icmlaffiliation{add2}{Key Laboratory of New Generation Artificial Intelligence Technology and Its Interdisciplinary Applications (Southeast University), Ministry of Education, China}

  \icmlcorrespondingauthor{Jing Wang}{wangjing91@seu.edu.cn}
  \icmlcorrespondingauthor{Xin Geng}{xgeng@seu.edu.cn}

  \icmlkeywords{Machine Learning, ICML}

  \vskip 0.3in
]



\printAffiliationsAndNotice{}  

\begin{abstract}
  The increasing scale and complexity of modern model parameters underscore the importance of pre-trained models. However, deployment often demands architectures of varying sizes, exposing limitations of conventional pre-training and fine-tuning. 
  To address this, we propose SWEET, a self-supervised framework that performs constraint-based pre-training to enable scalable initialization in vision tasks. 
  Instead of pre-training a fixed-size model, we learn a shared weight template and size-specific weight scalers under Tucker-based factorization, which promotes modularity and supports flexible adaptation to architectures with varying depths and widths. Target models are subsequently initialized by composing and reweighting the template through lightweight weight scalers, whose parameters can be efficiently learned from minimal training data.
  To further enhance flexibility in width expansion, we introduce width-wise stochastic scaling, which regularizes the template along width-related dimensions and encourages robust, width-invariant representations for improved cross-width generalization.  
  Extensive experiments on \textsc{classification}, \textsc{detection}, \textsc{segmentation} and \textsc{generation} tasks demonstrate the state-of-the-art performance of SWEET for initializing variable-sized vision models.
\end{abstract}

\vspace{-0.2in}
\section{Introduction}
With the rapid growth of model scale, training from scratch has become increasingly inefficient~\cite{liu2021swin, wu2021cvt}, making pre-training a cornerstone of modern visual learning, particularly in data-limited scenarios~\cite{qiu2020pre, han2021pre}.
However, traditional pre-training paradigms focus predominantly on maximizing performance on the pre-training dataset, producing models tightly coupled to a specific scale and downstream task domain (e.g., a ViT-L for image classification).

\begin{figure}
    \centering
    \includegraphics[width=\linewidth]{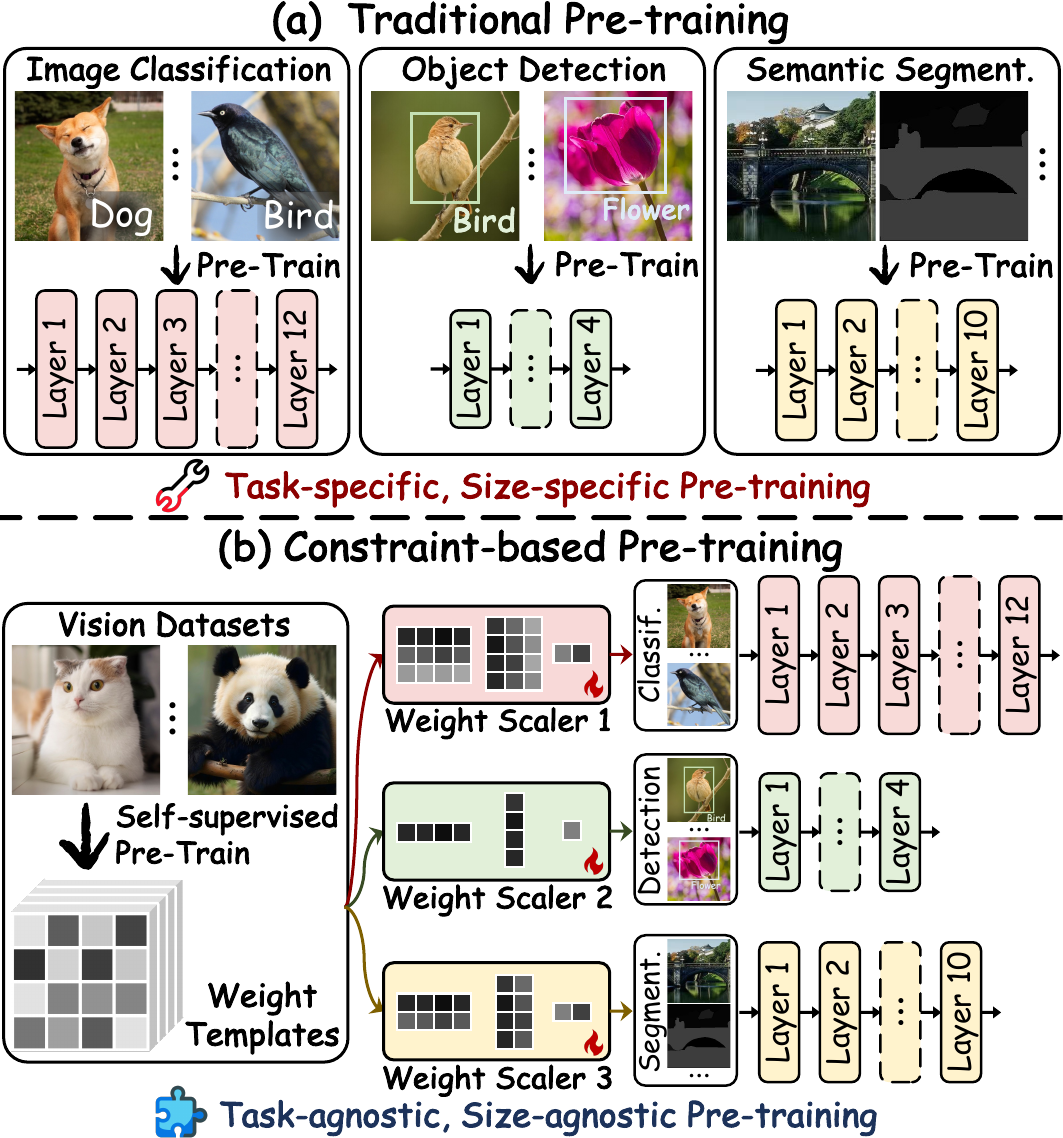}
    \vspace{-0.25in}
    \caption{(a) Traditional pre-training paradigms produce fixed-size, task-specific models, which are difficult to adapt to downstream architectures with varying scales and task requirements.
    (b) SWEET adopts a constraint-based pre-training paradigm that extracts weight templates under structured constraints in a self-supervised manner, enabling flexible cross-scale and cross-task model initialization and efficient knowledge transfer.}
    \label{fig:moti}
    \vspace{-0.25in}
\end{figure}

In practice, deployment is constrained by computational resources and downstream task requirements~\cite{zhang2022minivit}, necessitating models of diverse scales and task-specific capabilities.
Consequently, models that deviate from the configurations of off-the-shelf pre-trained models often necessitate retraining or knowledge distillation~\cite{gou2021knowledge}, imposing substantial computational overhead.
Although several methods initialize downstream models of different scales by reusing or transforming pre-trained weights for computational efficiency~\cite{lan2019albert, wang2022learngene, wang2023learning, xu2023initializing}, they often disrupt the structural coherence of pre-trained representations, leading to notable performance degradation.

Thus, recent advances shift pre-training from fixed-architecture optimization to decomposable parameter learning~\cite{xie2025kind, xie2025divcontrol}, enabling scalable downstream initialization.
A representative method, WAVE~\cite{feng2024wave}, formulates pre-training as a constraint-based optimization problem, training Vision Transformers under Kronecker-based constraints to learn structured \textit{\textbf{Weight Templates}} instead of full model parameters.
These templates are composed via the Kronecker product with lightweight \textit{\textbf{Weight Scalers}}, enabling scalable model initialization with negligible computational overhead.

\vspace{-0.03in}
Despite its initial success, WAVE exhibits limitations that constrain flexible and extensible model initialization.
Structurally, existing weight templates are restricted to homogeneous parameterizations, hindering parameter sharing across heterogeneous components such as attention heads and feed-forward modules. Furthermore, although they support width expansion, their fixed dimensionality limits adaptability to arbitrary widths.
Task-wise, while effective for image classification, weight templates demonstrate limited transferability to heterogeneous tasks such as semantic segmentation and object detection.
Together, these structural and task-related limitations prevent weight templates from serving as a universal initialization strategy for diverse visual models.

\vspace{-0.03in}
To address these limitations, we propose SWEET, a framework that learns self-supervised weight templates for scalable initialization across models of diverse sizes and visual tasks.
Specifically, to facilitate parameter sharing across heterogeneous components, SWEET reorganizes and concatenates all parameters from different layers and modules into a unified weight matrix $\mathcal{W}$.
Unlike prior constraint-based approaches~\cite{feng2024wave, xie2024fine}, SWEET reconstructs $\mathcal{W}$ via Tucker-based constraints~\cite{malik2018low} (see Eq.~\eqref{equ:tucker}), where $\mathcal{G}$ serves as the weight template and $(U, V, X)$ act as lightweight scalers, yielding a compact yet flexible parameter representation. 
A low-rank constraint is imposed on $\mathcal{G}$ as a ``bottleneck'' to condense size-agnostic knowledge~\cite{feng2023genes}.

\vspace{-0.03in}
To capture task-agnostic knowledge across diverse vision tasks, SWEET employs \textbf{\textit{a self-supervised pre-training objective}} to train weight templates, decoupling their representations from task-specific supervision and promoting generalizable visual features.
Furthermore, to enhance adaptability to downstream models of arbitrary widths, we apply width-wise stochastic scaling during pre-training via dropout~\cite{cai2020once} on the weight scalers, improving the robustness of weight templates across diverse width configurations.

\vspace{-0.03in}
SWEET substantially reduces computational cost compared to conventional full-model pre-training, as its structured constraints and low-rank bottleneck filter non-transferable knowledge, enabling faster convergence while preserving transferable representations for scalable initialization and cross-task adaptation.
Extensive experiments demonstrate SWEET’s state-of-the-art performance in initializing variable-sized models across diverse vision tasks.
On average across five model sizes, SWEET improves \textsc{Image classification} accuracy by 1.60\%.
Comparable advantages are observed on other vision tasks, with improvements of 2.04 AP, 2.76 mIoU, and 2.19 FID in \textsc{Object Detection}, \textsc{Semantic Segmentation}, and \textsc{Image Generation}, respectively.


Our contributions are as follows:
1)~We propose SWEET, a self-supervised framework for pre-training structured weight templates that enable scalable initialization across varying model scales and diverse vision tasks.
2)~We introduce width-wise stochastic scaling, a novel regularization strategy that enhances the robustness and adaptability of weight templates for initializing models with varying widths.
3)~We establish a comprehensive benchmark for multi-scale model initialization across vision tasks, demonstrating that SWEET consistently outperforms existing methods in both cross-scale and cross-task model initialization.

\vspace{-0.07in}
\section{Related Work}
\vspace{-0.02in}
\subsection{Model Initialization}
\vspace{-0.02in}
Model initialization is a fundamental factor influencing optimization efficiency and final performance~\cite{narkhede2022review, hanin2018start}. 
Early methods relied on hand-crafted heuristics for random initialization~\cite{glorot2010understanding, chen2021empirical}. 
With the rise of large-scale pre-training, initialization is now commonly inherited from pre-trained models, making fine-tuning the dominant paradigm~\cite{qiu2020pre, zhang2024vision}.

To better leverage fixed-architecture pre-trained models for initializing models of varying sizes, several methods explore scalable initialization strategies.
Mimetic Initialization~\cite{trockman2023mimetic} leverages parameter patterns identified in pre-trained models to initialize new ones, while GHN~\cite{knyazev2021parameter, knyazev2023canwescale} predict target model parameters using a graph hypernetwork conditioned on the computational graph. Weight Selection~\cite{xu2023initializing} directly transfers selected parameters from larger models to smaller ones.
Despite these advances, representations in conventionally pre-trained models are entangled within parameter matrices, making direct splitting or transformation prone to negative transfer due to parameter mismatches or feature disruption.
Our SWEET addresses this by pre-training structured weight templates rather than full models, enabling scalable initialization across model sizes.

\vspace{-0.05in}
\subsection{Learngene and Weight Templates}
\textsc{Learngene}~\cite{feng2023genes, wang2023learngene} is a biologically inspired knowledge transfer paradigm that encapsulates size-agnostic knowledge into modular neural units, termed learngenes, facilitating efficient adaptation across model scales.
Early learngene methods primarily operate on a layer-wise basis.
Heur-LG~\cite{wang2022learngene} identifies learngenes as layers with minimal gradient variation during continual learning, while Auto-LG~\cite{wang2023learngene} selects layers whose representations best align with the target network via meta-learning.
TLEG~\cite{xia2024transformer} models learngenes as pairs of base layers that can be linearly combined to initialize models of varying depths.

WAVE~\cite{feng2024wave} advances this line of work by introducing Weight Templates, overcoming layer-wise constraints by representing each weight matrix as a weighted combination of concatenated templates.
Our SWEET extend this paradigm with \textbf{\textit{self-supervised weight templates}}, substantially enhancing initialization flexibility and enabling universal model initialization across visual tasks.

\vspace{-0.07in}
\section{Methods}
\vspace{-0.04in}
\subsection{Preliminaries}
\vspace{-0.04in}
\subsubsection{Masked Autoencoders (MAE)} 
\vspace{-0.04in}
MAE~\cite{he2022masked} is a self-supervised pre-training framework based on ViTs that learns visual representations by reconstructing randomly masked image patches. 
Given an input image, a large portion of patches (typically 75\%) is masked, and the encoder processes only the visible patches to produce latent representations. 
A lightweight decoder then reconstructs the masked patches, optimized via
\begin{equation}
    \mathcal{L}_{\text{MAE}} = \frac{1}{|\mathcal{M}|}\sum_{i\in \mathcal{M}} \|\hat{x}_i - x_i\|_2^2
\label{equ:loss}
\end{equation}
where $\mathcal{M}$ is the set of masked patches, and $x_i$ and $\hat{x}_i$ denote the original and reconstructed patch embeddings.
This self-supervised objective encourages the encoder to extract fundamental, generalizable visual features that are broadly transferable across downstream vision tasks.

\vspace{-0.1in}
\subsubsection{Vision Transformer (ViT)} 
\vspace{-0.02in}
ViT~\cite{dosovitskiy2020image} comprises $L$ stacked layers, each containing a multi-head self-attention (MSA) followed by a multi-layer perceptron (MLP). 
In MSA, $h$ attention heads process the input, and their concatenated outputs are projected using a learnable weight matrix $W_o \in \mathbb{R}^{hd \times D}$:
\vspace{-0.02in}
\begin{equation}
    \text{MSA} = \text{concat}(A_1, A_2, ..., A_{h}) W_{o}\,,\;W_{o}\in \mathbb{R}^{hd\times D}
\label{equ:msa}
\end{equation}
Within a single attention head $A_i$, queries $Q_i$, keys $K_i$, and values $V_i \in \mathbb{R}^{N \times d}$ are obtained via learnable projections $W_q^i$, $W_k^i$, and $W_v^i \in \mathbb{R}^{D \times d}$, and self-attention is given by
\begin{equation}
    A_i = \text{softmax}\Big(\frac{Q_i K_i^\top}{\sqrt{d}}\Big)V_i, \quad A_i \in \mathbb{R}^{N \times d}
\end{equation}
where $N$ is the number of input patches, $D$ is the patch embedding dimension, and $d$ is the attention head dimension, typically $D=hd$ in standard multi-head self-attention.


MLP consists of two linear projections $W_{\text{in}}\in \mathbb{R}^{D\times D'}$ and $W_{\text{out}}\in \mathbb{R}^{D'\times D}$ with a GELU~\cite{hendrycks2016gaussian} activation, formulated as:
\begin{equation}
    \text{MLP}(x)=\text{GELU}(xW_{\text{in}}+b_1)W_{\text{out}} + b_2
\label{equ:mlp}
\end{equation}
where $b_1$, $b_2$ are bias and $D'$ is the hidden layer dimension, which is typically set to $D'=4D$ in standard ViT.

\begin{figure*}
    \centering
    \includegraphics[width=\linewidth]{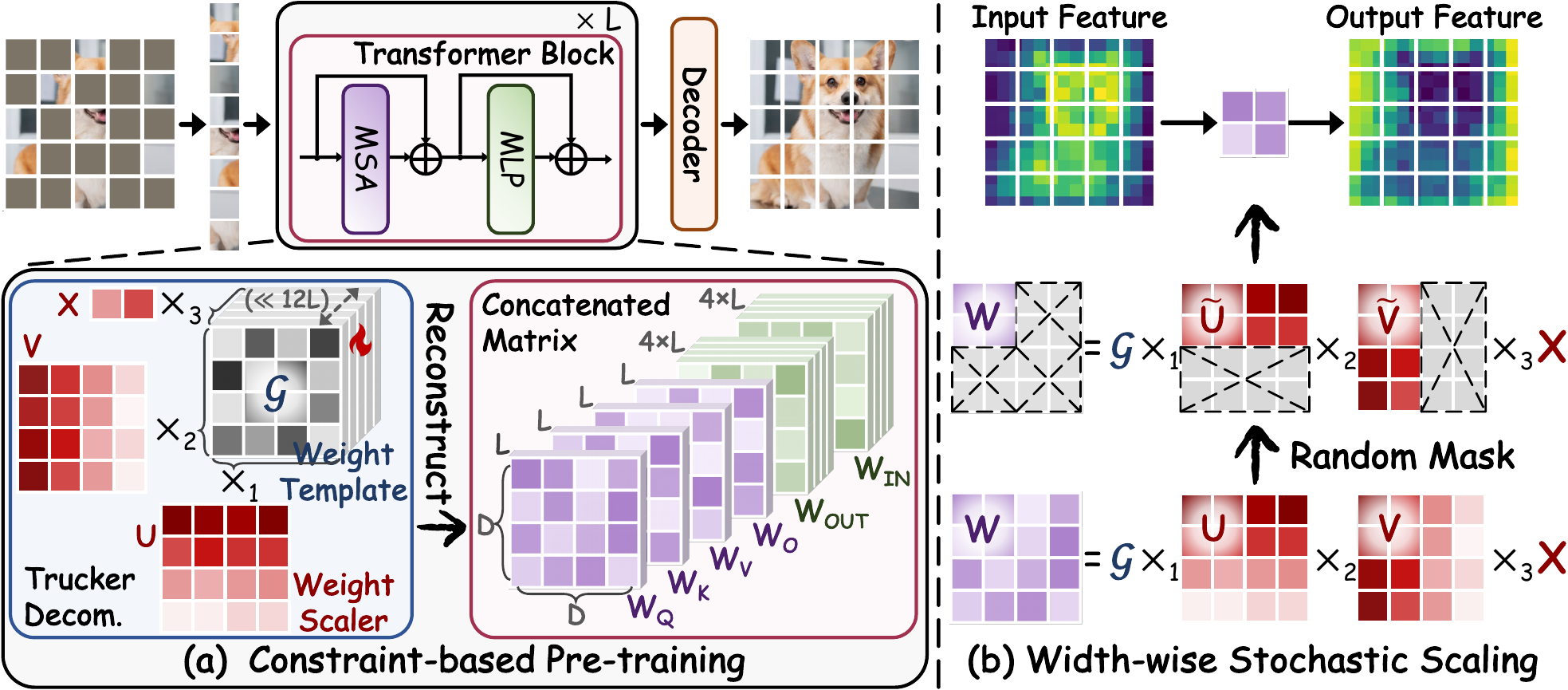}
    \vspace{-0.19in}
    \caption{\textbf{Overview of SWEET.} 
    \textbf{(a)~Constraint-based Pre-training} of the weight template $\mathcal{G}$ with weight scalers $(U, V, X)$, where Tucker- and low-rank constraints are applied to condense size-agnostic knowledge within the template.
    A self-supervised objective guides pre-training to learn generalizable visual representations, enhancing the template’s universality across diverse vision tasks.
    \textbf{(b)~Width-wise Stochastic Scaling} randomly masks weight scalers along the width during pre-training, discouraging overfitting to a specific width and promoting the organization of template knowledge for flexible adaptation across models of varying widths.}
    \label{fig:main}
    \vspace{-0.14in}
\end{figure*}

\vspace{-0.05in}
\subsection{Tucker-based Weight Template}
\vspace{-0.05in}
SWEET learns structured weight templates through a constraint-based pre-training paradigm, in which structured constraints are imposed to learn templates instead of a full model, enabling scalable and regularized initialization.
Representative works such as WAVE~\cite{feng2024wave} assign dedicated templates to individual components, focusing primarily on cross-scale knowledge sharing while overlooking shared patterns across components (e.g., between MSA and FFN), thereby limiting the generality of templates.

To address this limitation, we first aggregate the primary weight matrices of an $L$-layer ViT, $\theta = \{W_{q}^{(1\thicksim L)}$, $W_{k}^{(1\thicksim L)}$, $W_{v}^{(1\thicksim L)}$, $W_{o}^{(1\thicksim L)}$, $W_{\text{in}}^{(1\thicksim L)}$, $W_{\text{out}}^{(1\thicksim L)}\}$\footnote{$W_{q}^{(1\thicksim L)}$ denotes ${W_{q}^{(1)}, \dots, W_{q}^{(L)}}$ for brevity}, into a unified weight matrix $\mathcal{W} \in \mathbb{R}^{L \times P}$, where each row represents a layer and $P = D \cdot (4hd + 2D')=12 D \cdot D$, thereby bridging the boundaries between heterogeneous model components.

Next, we impose structured constraints on $\mathcal{W}$. Instead of Kronecker~\cite{feng2024wave} or SVD-based constraints~\cite{xie2024fine}, we adopt a more general Tucker-based constraint, of which Kronecker arise as special cases (see Appendix~A.2 for a brief proof).
Specifically, the unified weight matrix $\mathcal{W}$ is reconstructed as
\begin{equation}
    \mathcal{W} \Leftarrow \mathcal{G} \times_1 X \times_2 U \times_3 V,
\label{equ:tucker}
\end{equation}
Here, $\mathcal{G} \in \mathbb{R}^{r_1 \times r_2 \times r_3}$ is the core tensor, serving as a universal \textbf{\textit{Weight Template}} encoding size-agnostic knowledge, while the lightweight \textbf{\textit{Weight Scalers}} $(X, U, V)$ modulate its reuse and composition to reconstruct $\mathcal{W}$, with $X \in \mathbb{R}^{12L \times r_1}$, $U \in \mathbb{R}^{D \times r_2}$, $V \in \mathbb{R}^{D \times r_3}$. The operator $\times_i$ denotes mode-$i$ tensor-matrix multiplication.

To capture size-agnostic knowledge, we impose a low-rank constraint on the weight template $\mathcal{G}$, with $r_1 \times r_2 \times r_3 \ll L \times P$, following~\cite{feng2023genes, feng2024wave, xie2024fine}.
Acting as a bottleneck~\cite{zador2019critique}, this constraint concentrates reusable information within $\mathcal{G}$, filtering knowledge that poorly transfers across model sizes and thereby enhancing template transferability and pre-training efficiency.

\subsection{Self-Supervised Pre-Training and Width-Wise Stochastic Scaling}
\label{sec:wss}
Early weight template training methods~\cite{feng2024wave, xie2024fine} optimize conventional cross-entropy or reconstruction losses on image classification or generation, restricting their applicability to specific vision tasks and preventing their use as universal visual model initializers.

To capture generalizable visual knowledge beyond task-specific patterns, SWEET trains weight templates in a self-supervised manner.
Specifically, after aggregating the weights of an $L$-layer ViT as in Eq.~\eqref{equ:tucker}, SWEET jointly optimizes the weight template $\mathcal{G}$ and scalers $(X, U, V)$, through which the full model parameters are implicitly reconstructed. 
Formally, the pre-training objective is
\begin{equation}
\begin{aligned}
    \min_{\mathcal{G}, U, V, X} \;\;& 
    \mathcal{L}_{\text{MAE}}\big(f_{\theta}(x)\big), \\
    \text{s.t.} \;\;& 
    \mathrm{concat}(\theta) = \mathcal{W} = \mathcal{G} \times_1 X \times_2 U \times_3 V.
\end{aligned}
\label{equ:ob}
\end{equation}
where $\mathcal{L}_{\text{MAE}}$ is the reconstruction loss defined in Eq.~\eqref{equ:loss}. 
This indirect optimization decouples knowledge extraction from specific parameter values, regularizes the initialization space, and promotes the learning of universal and structural visual patterns (see Algorithm~1).

Traditional layer-based learngene methods~\cite{wang2022learngene, wang2023learngene, xia2024transformer} struggle to generalize across model widths, as their transfer units are intrinsically tied to fixed layers, while WAVE is constrained by the fixed dimensionality of its weight templates.
To support flexible width expansion, SWEET introduces \textbf{\textit{width-wise stochastic scaling}}, which applies structured dropout to the weight scalers, encouraging the weight template to capture width-robust knowledge:
\begin{equation}
    \tilde{U} = M_U \odot U, \quad \tilde{V} = M_V \odot V
\end{equation}
where $M_U$ and $M_V$ are independently sampled binary masks, drawn from a predefined distribution over width configurations, and $\odot$ denotes element-wise multiplication.
The reconstructed weight matrix is
\begin{equation}
    \mathcal{W} \Leftarrow \mathcal{G} \times_1 X \times_2 \tilde{U} \times_3 \tilde{V}.
\end{equation}
By introducing stochastic scaling during pre-training, the model is prevented from overfitting to a fixed width, forcing it to reorganize knowledge along the width dimension. 
This reinforces width-invariant structural representations in the low-index dimensions of the weight template, enabling stable adaptation to models with varying widths.

Following~\cite{yao2025reconstruction, li2025back}, to enhance generalization, we incorporate several architectural enhancements, including SwiGLU~\cite{shazeer2020glu}, RMSNorm~\cite{zhang2019root}, and RoPE~\cite{su2024roformer}, originally proposed for language models.

\subsection{Scalable Model Initialization}
Benefiting from structured constraints and a low-rank bottleneck that filters non-transferable knowledge, constraint-based pre-training is substantially efficient than conventional full-model pre-training and incurs a once-for-all cost (see App.~A.1 for a brief theoretical analysis).
The learned weight template $\mathcal{G}$ enables zero- or negligible-cost initialization of models with arbitrary sizes. During initialization, the template is kept frozen, while the lightweight scalers $(X, U, V)$ are directly selected or randomly initialized to match the target scale and optionally optimized, enabling flexible and scalable model instantiation.

Specifically, given a downstream model with parameters $\theta_{\star}$, we form a unified weight matrix $\mathcal{W}_{\star} = \mathrm{concat}(\theta_{\star}) \in \mathbb{R}^{L_{\star} \times P_{\star}}$ by concatenating its layer-wise parameters, where $L_{\star}$ is the number of layers and $P_{\star}$ denotes the total per-layer width of the concatenated weight matrices.
The weight scalers are \textbf{\textit{randomly initialized}} or  \textbf{\textit{directly inherited}} from pre-trained $(X, U, V)$ to fit the target dimensions, producing $X_{\star} \in \mathbb{R}^{12L_{\star} \times r_1}$, $U_{\star} \in \mathbb{R}^{D_{\star} \times r_2}$, and
$V_{\star} \in \mathbb{R}^{D_{\star} \times r_3}$ as above, which reconstruct $\mathcal{W}_{\star}$ while retaining the size-agnostic knowledge embedded in the frozen template $\mathcal{G}$.

Benefiting from width-wise stochastic scaling, target models with modest scale variations can be effectively initialized by directly selecting dimension-aligned slices from pre-trained $(X, U, V)$.
For extremely compact models, template knowledge can be further adapted by lightly optimizing the scalers on a small dataset, while keeping the template $\mathcal{G}$ frozen:
\vspace{-0.03in}
\begin{equation} 
\begin{aligned} 
    \min_{U_{\star}, V_{\star}, X_{\star}} \;\; & \mathcal{L}_{\text{MAE}}\big(f_{\theta_{\star}}(x)\big), \\ \text{s.t.} \;\;& \mathrm{concat}(\theta_{\star}) = \mathcal{W}_{\star} = \mathcal{G} \times_1 X_{\star} \times_2 U_{\star} \times_3 V_{\star}. 
\end{aligned} 
\label{equ:train_S} 
\end{equation}
Owing to the small size of the weight scalers—typically only a few thousand parameters—optimization converges within a few hundred iterations ($\approx 0.16$ epoch), imposing negligible computational overhead. Once learned, the target model is initialized via Eq.~\eqref{equ:tucker}, after which it can be trained conventionally without additional constraints.

\section{Experiments}
\subsection{Experimental Setup}
\vspace{-0.03in}
\paragraph{Vision Tasks and Datasets}
SWEET is first pre-trained in a self-supervised manner on ImageNet-1K~\cite{deng2009imagenet} to learn the weight template, and is subsequently evaluated on four representative vision tasks: \textsc{Image Classification} and \textsc{Image Generation} on ImageNet-1K, \textsc{Semantic Segmentation} on ADE20K~\cite{zhou2019semantic}, and \textsc{Object Detection} on COCO~\cite{lin2014microsoft}.
Additional details are provided in Appendix~B.2.

\vspace{-0.08in}
\paragraph{Network Structures}
We adopt ViT-Base (ViT-B/16)~\cite{dosovitskiy2020image} as the backbone for weight template pre-training. 
To evaluate SWEET’s scalable initialization across depth and width, we consider ViT configurations with depths $L \in \{3, 6, 12\}$ and widths adjusted via the number of attention heads, $H \in \{3, 6, 12\}$.

\vspace{-0.08in}
\paragraph{Evaluation Metrics}
We adopt standard metrics for each vision task. \textsc{Image Classification} is evaluated using Top-1 and Top-5 accuracy. \textsc{Image Generation} is assessed via Fréchet Inception Distance (FID)~\cite{heusel2017gans} and Inception Score (IS)~\cite{salimans2016improved} to capture visual fidelity and diversity. \textsc{Object Detection} is measured by mean Average Precision for bounding boxes and masks ($\text{AP}^\text{box}$, $\text{AP}^\text{mask}$), while \textsc{Semantic Segmentation} is quantified by mean Intersection over Union (mIoU) and mean pixel accuracy (mAcc).

\vspace{-0.14in}
\paragraph{Training Details}
The weight template is pre-trained in a self-supervised manner for 450 epochs on a batch size of 1024, using AdamW with a learning rate of $6 \times 10^{-4}$ and a cosine learning rate scheduler on an NVIDIA RTX 4090 GPU. 
In comparison, baseline methods use the official MAE weights from~\cite{he2022masked}, traditionally pre-trained for 800 epochs with a batch size of 4096, making SWEET significantly more computationally efficient. Additional details are provided in Appendix~B.3.

\vspace{-0.05in}
\subsection{Baselines}
\vspace{-0.03in}
We compare SWEET with state-of-the-art scalable model initialization methods.
1)~WT-Select~\cite{xu2023initializing} initializes target models by directly selecting and reusing weight subsets from a pre-trained model according to predefined rules. 
2)~DMAE~\cite{bai2023masked} distills knowledge from a teacher model by training student models to reconstruct masked inputs and align their intermediate feature maps.
3)~Isomorphic Pruning~\cite{fang2024isomorphic} partitions parameters based on computational topology and ranks them within groups to guide pruning. 
4)~WAVE~\cite{feng2024wave}, a representative constraint-based method, constructs weight templates via Kronecker-based constraints.

\renewcommand{\arraystretch}{0.97}
\vspace{-0.05in}
\section{Results}
\vspace{-0.03in}
\subsection{Performance on Discriminative Vision Tasks}
\vspace{-0.03in}
\begin{table*}[!t]
    \centering
    \setlength{\tabcolsep}{0.6 mm} 
    \caption{Scalable initialization performance of SWEET across model scales on fundamental vision tasks.
    $L_l H_h$ denotes models with $l$ layers and $h$ attention heads, corresponding to the model's depth and width, respectively.
    ``Para.(M)'' and ``FLOPs (G)'' indicate the parameter count and computational complexity for each model scale.
    Models for \textsc{Image Classification} and \textsc{Object Detection} are trained for 30 epochs, while \textsc{Semantic Segmentation} are trained for 160K iterations after initialization.}
    \vspace{-0.1in}
    \resizebox{\linewidth}{!}{
        \begin{tabular}{@{}lcc|cc|cc|cc|cc|>{\columncolor{gray!10}}c>{\columncolor{gray!10}}c@{}}
        \toprule[1.5pt]
         & \multicolumn{2}{c}{$\textcolor{red!40}{\boldsymbol{L_3}} \textcolor{blue!100}{\boldsymbol{H_{12}}}$} & \multicolumn{2}{c}{$\textcolor{red!65}{\boldsymbol{L_6}} \textcolor{blue!100}{\boldsymbol{H_{12}}}$} & \multicolumn{2}{c}{$\textcolor{red!65}{\boldsymbol{L_6}} \textcolor{blue!65}{\boldsymbol{H_{6}}}$} & \multicolumn{2}{c}{$\textcolor{red!100}{\boldsymbol{L_{12}}} \textcolor{blue!65}{\boldsymbol{H_{6}}}$} & \multicolumn{2}{c|}{$\textcolor{red!100}{\boldsymbol{L_{12}}} \textcolor{blue!40}{\boldsymbol{H_{3}}}$} & \multicolumn{2}{c}{\multirow{2}{*}{Average}}   \\
         \cmidrule{2-11}
         Para./FLOPs & \multicolumn{2}{c}{\cellcolor{gray!15}{22.8{\small M} / 8.6{\small G}}} & \multicolumn{2}{c}{\cellcolor{gray!15}{44.0{\small M} / 17.0{\small G}}} & \multicolumn{2}{c}{\cellcolor{gray!15}{11.4{\small M} / 4.3{\small G}}} & \multicolumn{2}{c}{\cellcolor{gray!15}{22.1{\small M} / 8.5{\small G}}} & \multicolumn{2}{c|}{\cellcolor{gray!15}{5.7{\small M} / 2.2{\small G}}} \\
        \midrule[1.1pt]
        \midrule[1.1pt]
        \textsc{Image Classification} & Top1 & Top5 & Top1 & Top5 & Top1 & Top5 & Top1 & Top5 & Top1 & Top5 & Top1 & Top5 \\
        \midrule
        WT-Select~\cite{xu2023initializing} 
        & 66.44 & 86.46 
        & 75.91 & 92.42 
        & 67.03 & 86.93 
        & 66.51 & 86.81 
        & 43.49 & 68.58
        & \textit{63.87} & \textit{84.24}\\
        DMAE~\cite{bai2023masked} & 66.67 & 86.71 & 73.95 & 91.17 & 64.86
                & 85.65 & 65.70 & 86.17 & 49.07 & 73.77
                & \textit{64.05} & \textit{84.69}\\
        Iso. Pruning~\cite{fang2024isomorphic} 
        & 66.46 & 86.35 & 75.85 & 92.54 & 68.25
        & 88.00 & 70.86 & 89.67 & 44.53 & 68.89 
        & \textit{65.19} & \textit{85.09} \\
        WAVE~\cite{feng2024wave} 
        & 63.75 & 84.91 & 76.97 & 93.19 & 68.38 
        & 88.16 & 71.27 & 90.01 & 55.56 & 79.35
        & \textit{67.19} & \textit{87.12}\\
        \cellcolor{blue!12}{SWEET} 
               & \cellcolor{blue!12}{\textbf{67.21}} & \cellcolor{blue!12}{\textbf{86.93}} & \cellcolor{blue!12}{\textbf{77.42}} & \cellcolor{blue!12}{\textbf{93.25}} & \cellcolor{blue!12}{\textbf{70.34}}
               & \cellcolor{blue!12}{\textbf{89.22}} & \cellcolor{blue!12}{\textbf{71.70}} & \cellcolor{blue!12}{\textbf{90.24}} & \cellcolor{blue!12}{\textbf{57.28}} & \cellcolor{blue!12}{\textbf{80.31}}
               & \cellcolor{blue!12}{\textbf{\textit{68.79}}} & \cellcolor{blue!12}{\textbf{\textit{87.99}}} \\
        & \textcolor{mygreen}{$\uparrow$0.54} 
        & \textcolor{mygreen}{$\uparrow$0.22} 
        & \textcolor{mygreen}{$\uparrow$0.45} 
        & \textcolor{mygreen}{$\uparrow$0.06} 
        & \textcolor{mygreen}{$\uparrow$1.96}
        & \textcolor{mygreen}{$\uparrow$1.06} 
        & \textcolor{mygreen}{$\uparrow$0.42} 
        & \textcolor{mygreen}{$\uparrow$0.23} 
        & \textcolor{mygreen}{$\uparrow$1.72} 
        & \textcolor{mygreen}{$\uparrow$0.96}
        & \textcolor{mygreen}{$\uparrow$\textit{1.60}} 
        & \textcolor{mygreen}{$\uparrow$\textit{0.87}}\\
        \midrule[1.1pt]
        \midrule[1.1pt]
        \textsc{Object Detection} & $\text{AP}^\text{box}$ & $\text{AP}^\text{mask}$ & $\text{AP}^\text{box}$ & $\text{AP}^\text{mask}$ & $\text{AP}^\text{box}$ & $\text{AP}^\text{mask}$ & $\text{AP}^\text{box}$ & $\text{AP}^\text{mask}$ & $\text{AP}^\text{box}$ & $\text{AP}^\text{mask}$ & $\text{AP}^\text{box}$ & $\text{AP}^\text{mask}$ \\
        \midrule
        WT-Select~\cite{xu2023initializing}
            & 27.02 & 25.89 & 36.53 & 33.91 & 25.02
            & 23.89 & 34.09 & 31.40 & 23.72 & 22.66
            & \textit{29.28} & \textit{27.55} \\
        DMAE~\cite{bai2023masked} 
            & 25.55 & 24.29 & 29.85 & 27.80 & 25.16
            & 23.88 & 32.82 & 30.20 & 26.51 & 24.95
            & \textit{27.98} & \textit{26.22} \\
        Iso. Pruning~\cite{fang2024isomorphic} 
            & 26.97 & 25.75 & 36.64 & 33.94 & 23.59
            & 22.54 & 35.74 & 32.85 & 23.13 & 21.96 
            & \textit{29.21} & \textit{27.41}\\
        WAVE~\cite{feng2024wave}   
            & 27.94 & 26.63 & 34.19 & 31.91 & 26.14 
            & 24.93 & 33.55 & 31.15 & 25.94 & 24.45
            & \textit{29.55} & \textit{27.81} \\
        \cellcolor{blue!12}{SWEET} 
               & \cellcolor{blue!12}{\textbf{28.88}} & \cellcolor{blue!12}{\textbf{27.23}} & \cellcolor{blue!12}{\textbf{38.16}} & \cellcolor{blue!12}{\textbf{35.16}} & \cellcolor{blue!12}{\textbf{27.33}}
               & \cellcolor{blue!12}{\textbf{25.80}} & \cellcolor{blue!12}{\textbf{36.13}} & \cellcolor{blue!12}{\textbf{33.02}} & \cellcolor{blue!12}{\textbf{27.46}} & \cellcolor{blue!12}{\textbf{25.82}} &
               \cellcolor{blue!12}{\textbf{\textit{31.59}}} &
               \cellcolor{blue!12}{\textbf{\textit{29.41}}}\\
        & \textcolor{mygreen}{$\uparrow$0.94} 
        & \textcolor{mygreen}{$\uparrow$0.60} 
        & \textcolor{mygreen}{$\uparrow$1.53} 
        & \textcolor{mygreen}{$\uparrow$1.22} 
        & \textcolor{mygreen}{$\uparrow$1.20}
        & \textcolor{mygreen}{$\uparrow$0.87} 
        & \textcolor{mygreen}{$\uparrow$0.39} 
        & \textcolor{mygreen}{$\uparrow$0.17} 
        & \textcolor{mygreen}{$\uparrow$0.95} 
        & \textcolor{mygreen}{$\uparrow$0.87}
        & \textcolor{mygreen}{$\uparrow$\textit{2.04}} 
        & \textcolor{mygreen}{$\uparrow$\textit{1.59}}\\
        \midrule[1.1pt]
        \midrule[1.1pt]
        \textsc{Semantic Segmentation} & mIoU & mAcc & mIoU & mAcc & mIoU & mAcc & mIoU & mAcc & mIoU & mAcc & mIoU & mAcc \\
        \midrule
        WT-Select~\cite{xu2023initializing} 
            & 28.24 & 36.51 & 35.78 & 44.75 & 26.36
            & 34.83 & 29.89 & 38.70 & 22.29 & 30.22
            & \textit{28.51} & \textit{37.00} \\
        DMAE~\cite{bai2023masked} 
            & 27.82 & 36.38 & 32.64 & 41.53 & 28.07
            & 37.35 & 31.43 & 40.92 & 24.79 & 33.49
            & \textit{28.95} & \textit{37.93} \\
        Iso. Pruning~\cite{fang2024isomorphic} 
            & 27.34 & 35.00 & 35.89 & 44.73 & 26.89
            & 35.12 & 31.44 & 40.05 & 24.12 & 32.00 
            & \textit{29.14} & \textit{37.38} \\
        WAVE~\cite{feng2024wave} 
            & 28.03 & 36.23 & 33.84 & 42.26 & 29.15 
            & 37.82 & 32.55 & 41.39 & 28.04 & 37.20
            & \textit{30.32} & \textit{38.98} \\
        \cellcolor{blue!12}{SWEET} 
               & \cellcolor{blue!12}{\textbf{29.35}} & \cellcolor{blue!12}{\textbf{37.93}} & \cellcolor{blue!12}{\textbf{38.39}} & \cellcolor{blue!12}{\textbf{47.81}} & \cellcolor{blue!12}{\textbf{31.46}}
               & \cellcolor{blue!12}{\textbf{40.78}} & \cellcolor{blue!12}{\textbf{37.32}} & \cellcolor{blue!12}{\textbf{47.29}} & \cellcolor{blue!12}{\textbf{28.88}} & \cellcolor{blue!12}{\textbf{38.26}} &
               \cellcolor{blue!12}{\textbf{\textit{33.08}}} &
               \cellcolor{blue!12}{\textbf{\textit{42.41}}}\\
        & \textcolor{mygreen}{$\uparrow$1.11} 
        & \textcolor{mygreen}{$\uparrow$1.42} 
        & \textcolor{mygreen}{$\uparrow$2.50} 
        & \textcolor{mygreen}{$\uparrow$3.06} 
        & \textcolor{mygreen}{$\uparrow$2.31}
        & \textcolor{mygreen}{$\uparrow$2.96} 
        & \textcolor{mygreen}{$\uparrow$4.77} 
        & \textcolor{mygreen}{$\uparrow$5.90} 
        & \textcolor{mygreen}{$\uparrow$0.84} 
        & \textcolor{mygreen}{$\uparrow$1.06}
        & \textcolor{mygreen}{$\uparrow$\textit{2.76}} 
        & \textcolor{mygreen}{$\uparrow$\textit{3.43}}\\
               
        \bottomrule[1.5pt]
        \end{tabular}   
    }    
    \label{tab:basic_vision}
    \vspace{-0.15in}
\end{table*}

\begin{figure*}[!t]
    \centering
    \includegraphics[width=\linewidth]{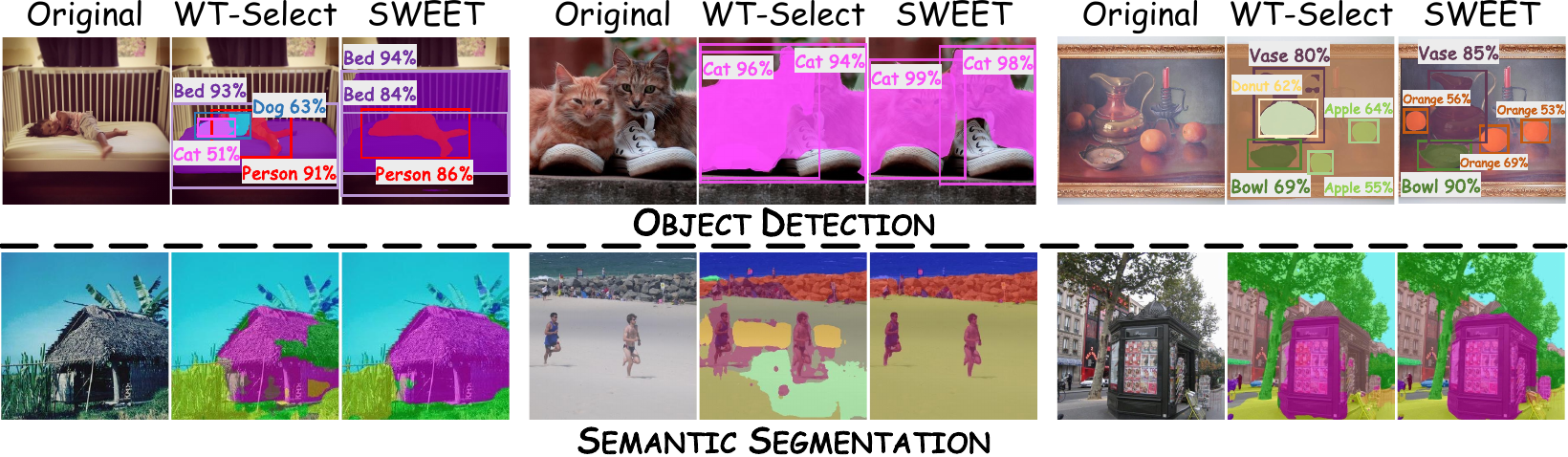}
    \vspace{-0.2in}
    \caption{Selected visualizations of \textsc{Object Detection} and \textsc{Semantic Segmentation} for SWEET-initialized models.}
    \label{fig:basic_vision}
    \vspace{-0.23in}
\end{figure*}

\subsubsection{Image Classification}
\vspace{-0.03in}
\label{sec:classification}
We first evaluate SWEET on \textsc{Image Classification}, a standard vision task whose accuracy is highly sensitive to low- and mid-level features, making it a reliable measure of initialization quality.
As shown in Table~\ref{tab:basic_vision}, SWEET consistently improves Top-1 accuracy across all model configurations, averaging 1.60 above the strongest baseline.
This indicates that the learned weight templates capture multi-scale, category-discriminative features—such as edges, textures, and local shape patterns—that generalize across model depths and widths.

Beyond accuracy, SWEET provides scalable initialization with negligible computational cost. Unlike distillation (e.g., DMAE) or pruning (e.g., Iso. Pruning), which require size-specific optimization or iterative fine-tuning, SWEET leverages pre-trained weight templates through a one-time lightweight weight scaler adaptation, efficiently supporting models of arbitrary sizes.
Notably, SWEET surpasses WAVE on image classification, even though WAVE is trained specifically for the task. This demonstrates that self-supervised weight templates effectively capture transferable visual features, and that sharing templates across all components with Tucker decomposition provides a more flexible and scalable model initialization.



\begin{table*}[!t]
    \centering
    \setlength{\tabcolsep}{0.6 mm} 
        \caption{Scalable initialization performance on \textsc{Image Generation}. Models are trained for 50 epochs after initialization.}  
        \vspace{-0.1in}
        \resizebox{\textwidth}{!}{
        \begin{tabular}{@{}lcc|cc|cc|cc|cc|>{\columncolor{gray!10}}c>{\columncolor{gray!10}}c@{}}
        \toprule[1.5pt]
        & \multicolumn{2}{c}{$\textcolor{red!40}{\boldsymbol{L_3}} \textcolor{blue!100}{\boldsymbol{H_{12}}}$} & \multicolumn{2}{c}{$\textcolor{red!65}{\boldsymbol{L_6}} \textcolor{blue!100}{\boldsymbol{H_{12}}}$} & \multicolumn{2}{c}{$\textcolor{red!65}{\boldsymbol{L_6}} \textcolor{blue!65}{\boldsymbol{H_{6}}}$} & \multicolumn{2}{c}{$\textcolor{red!100}{\boldsymbol{L_{12}}} \textcolor{blue!65}{\boldsymbol{H_{6}}}$} & \multicolumn{2}{c|}{$\textcolor{red!100}{\boldsymbol{L_{12}}} \textcolor{blue!40}{\boldsymbol{H_{3}}}$} & \multicolumn{2}{c}{\multirow{2}{*}{Average}} \\
        \cmidrule{2-11}
        Para./FLOPs & \multicolumn{2}{c}{\cellcolor{gray!15}{33.6{\small M} / 11.9{\small G}}} & \multicolumn{2}{c}{\cellcolor{gray!15}{64.3{\small M} / 23.8{\small G}}} & \multicolumn{2}{c}{\cellcolor{gray!15}{16.4{\small M} / 6.0{\small G}}} & \multicolumn{2}{c}{\cellcolor{gray!15}{31.8{\small M} / 11.9{\small G}}} & \multicolumn{2}{c|}{\cellcolor{gray!15}{8.1{\small M} / 3.0{\small G}}} \\
        \midrule[1.1pt]
        \midrule[1.1pt]
        \textsc{Image Generation} & FID & IS & FID & IS & FID & IS & FID & IS & FID & IS & FID & IS \\
        \midrule
        WT-Select~\cite{xu2023initializing} 
            & 31.64 & 21.42 & 17.47 & 33.61 & 35.09
            & 18.91 & 21.68 & 27.50 & 47.31 & 13.86 
            & \textit{30.64} & \textit{23.06} \\
        DMAE~\cite{bai2023masked} 
            & 29.09 & 23.18 & 15.50 & 37.30 & 33.50
            & 19.47 & 21.99 & 26.68 & 44.23 & 14.96
            & \textit{28.86} & \textit{24.32} \\
        Iso. Pruning~\cite{fang2024isomorphic} 
            & 32.29 & 21.23 & 18.68 & 31.53 & 37.60
            & 17.44 & 23.83 & 24.92 & 47.06 & 14.09 
            & \textit{31.89} & \textit{21.84} \\
        WAVE~\cite{feng2024wave} 
            & 31.54 & 21.30 & 16.61 & 34.37 & 34.58 
            & 18.85 & 20.37 & 28.17 & 45.35 & 14.66
            & \textit{29.69} & \textit{23.47} \\
        \cellcolor{blue!12}{SWEET} 
               & \cellcolor{blue!12}{\textbf{27.60}} & \cellcolor{blue!12}{\textbf{24.04}} & \cellcolor{blue!12}{\textbf{14.41}} & \cellcolor{blue!12}{\textbf{38.86}} & \cellcolor{blue!12}{\textbf{31.10}}
               & \cellcolor{blue!12}{\textbf{20.91}} & \cellcolor{blue!12}{\textbf{19.04}} & \cellcolor{blue!12}{\textbf{30.97}} & \cellcolor{blue!12}{\textbf{41.20}} & \cellcolor{blue!12}{\textbf{15.58}} & \cellcolor{blue!12}{\textbf{\textit{26.67}}} &
               \cellcolor{blue!12}{\textbf{\textit{26.07}}}\\
        & \textcolor{mygreen}{$\downarrow$1.49} 
        & \textcolor{mygreen}{$\uparrow$0.87} 
        & \textcolor{mygreen}{$\downarrow$1.09} 
        & \textcolor{mygreen}{$\uparrow$1.57} 
        & \textcolor{mygreen}{$\downarrow$2.40}
        & \textcolor{mygreen}{$\uparrow$1.44} 
        & \textcolor{mygreen}{$\downarrow$1.33} 
        & \textcolor{mygreen}{$\uparrow$2.80} 
        & \textcolor{mygreen}{$\downarrow$3.03} 
        & \textcolor{mygreen}{$\uparrow$0.63}
        & \textcolor{mygreen}{$\downarrow$\textit{2.19}} 
        & \textcolor{mygreen}{$\uparrow$\textit{1.76}} \\
        \bottomrule[1.5pt]
        \end{tabular}
        }
    \label{fig:generation}
\vspace{-0.1in}
\end{table*}

\begin{table*}[!t]
    \centering
    \setlength{\tabcolsep}{1.0mm}
    \caption{Performance of models (i.e., $\textcolor{red!65}{\boldsymbol{L_6}} \textcolor{blue!65}{\boldsymbol{H_{6}}}$) on \textsc{Image Classification} with downstream datasets measured by Top-1 Accuracy.}
    \vspace{-0.08in}
    \resizebox{\textwidth}{!}{
        \begin{tabular}{@{}lccccccc>{\columncolor{gray!15}}c@{}}
        \toprule[1.5pt]
             & {\small Oxford Flower} & {\small CUB-200} & {\small Stanford Cars} & {\small CIFAR10} & {\small CIFAR100} & {\small Food101} & {\small iNat-2019} & \textit{Average} \\
             \midrule
             WT-Select~\cite{xu2023initializing} & 76.65 & 54.99 & 60.98 & 95.21 & 74.10 & 81.95 & 62.67 & \textit{72.36} \\
             DMAE~\cite{bai2023masked} & \textbf{83.18} & 60.84 & 71.41 & 95.82 & 77.36 & 80.78 & 58.84 & \textit{75.46} \\
             Iso. Pruning~\cite{fang2024isomorphic} & 75.77 & 51.71 & 52.23 & 94.93 & 74.10 & 81.27 & 63.30 & \textit{70.47} \\
             WAVE~\cite{feng2024wave} & 80.26 & 56.77 & 57.71 & 93.71 & 75.58 & 82.42 & 63.70 & \textit{72.56} \\
             \cellcolor{blue!12}{SWEET} & \cellcolor{blue!12}{83.12} & \cellcolor{blue!12}{\textbf{61.51}} & \cellcolor{blue!12}{\textbf{81.03}} & \cellcolor{blue!12}{\textbf{97.03}} & \cellcolor{blue!12}{\textbf{79.36}} & \cellcolor{blue!12}{\textbf{82.50}} & \cellcolor{blue!12}{\textbf{64.22}} & {\textbf{\textit{78.40}}} \\
             & \textcolor{gray}{$\downarrow$0.06} & \textcolor{mygreen}{$\uparrow$0.67} & \textcolor{mygreen}{$\uparrow$9.63} & \textcolor{mygreen}{$\uparrow$1.21} & \textcolor{mygreen}{$\uparrow$2.00} & \textcolor{mygreen}{$\uparrow$0.08} & \textcolor{mygreen}{$\uparrow$0.53} & \textcolor{mygreen}{$\uparrow$2.01}\\
             \bottomrule[1.5pt]
        \end{tabular}
        }
    \label{tab:downstream}
\vspace{-0.17in}
\end{table*}

\vspace{-0.02in}
\subsubsection{Object Detection}
\vspace{-0.01in}
\label{sec:detection}
\textsc{Object Detection} demands more from initialization due to its reliance on accurate localization and semantic understanding. 
Despite this, SWEET consistently surpasses scalable initialization baselines across model scales, with stable gains of 2.04 and 1.59 in $\text{AP}^\text{box}$ and $\text{AP}^\text{mask}$, demonstrating its advantage in providing effective, size-agnostic initialization for complex vision tasks.

Unlike classification, object detection requires coordinated representations across the backbone and detection heads. By reconstructing models from a shared weight template, SWEET preserves inter-layer feature consistency, avoiding the misalignment and feature disruption introduced by direct weight selection and enabling reliable scaling without performance degradation. 
As shown in Fig.~\ref{fig:basic_vision}a, detectors initialized with SWEET achieve more accurate bounding box alignment and reduced localization errors, particularly for small and medium objects, indicating that the learned weight templates preserve spatial priors and hierarchical feature structures critical for localization.

\subsubsection{Semantic Segmentation}
\label{sec:segmentation}
Semantic segmentation imposes additional challenges for scalable initialization, as it requires dense, pixel-level predictions. Across varying model depths and widths, SWEET consistently outperforms baselines, achieving improvements of 2.76 and 3.43 in mIoU and mAcc.

\vspace{-0.02in}
Fig.~\ref{fig:basic_vision}b further validates this advantage. SWEET-initialized models produce more coherent segmentation maps with sharper boundaries and fewer fragmented regions, indicating that the learned weight templates effectively preserve fine-grained spatial structures and cross-layer consistency.

\vspace{-0.02in}
Collectively, these results show that SWEET provides a unified, scale-robust initialization across vision tasks by leveraging self-supervised weight templates that capture transferable features and preserve structural integrity. 
In contrast, WAVE, trained only on classification, shows weaker transferability to detection and segmentation, suggesting the benefit of self-supervised templates for cross-task initialization.

\vspace{-0.05in}
\subsection{Performance on Generative Vision Tasks}
\vspace{-0.04in}
\begin{figure*}[!t]
    \centering
    \includegraphics[width=\linewidth]{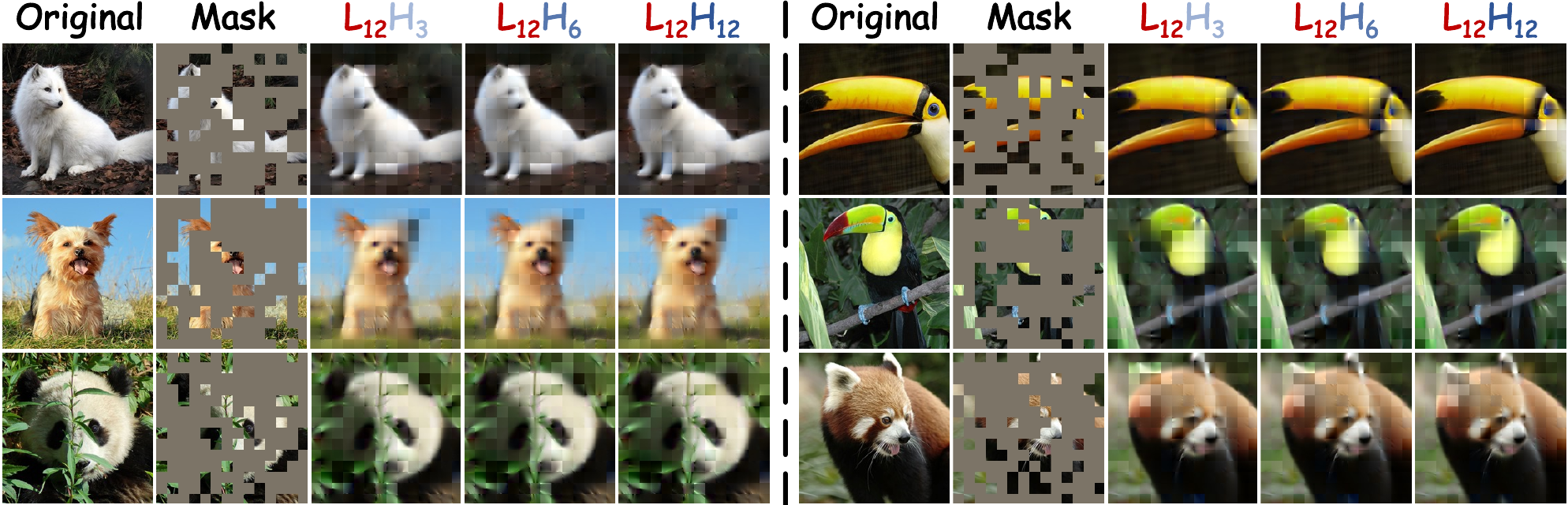}
    \vspace{-0.25in}
    \caption{Reconstructions of ImageNet validation images using SWEET pre-trained weight templates with a masking ratio of 75\%.}
    \label{fig:mae}
    \vspace{-0.15in}
\end{figure*}

Recent studies~\cite{yu2025representation, yun2025no, lei2025advancing} have demonstrated that representations learned from discriminative tasks can be leveraged to improve the efficiency of image generation.
Building on this, we evaluate self-supervised weight templates on \textsc{Image Generation}, a fundamentally more demanding task due to its requirement for holistic content synthesis.

\vspace{-0.03in}
Table~\ref{fig:generation} shows that SWEET consistently achieves lower FID ($\downarrow$2.19) and higher IS ($\uparrow$1.76) than all baselines, reflecting improved visual fidelity and diversity.
This advantage arises from weight templates, which preserve rich multi-scale visual priors through regularized constraints, whereas WT-Select and Iso. Pruning, by discarding parameters, can disrupt these priors and compromise cross-scale feature consistency.
Moreover, SWEET preserves priors more effectively than classification-trained templates , demonstrating the superior generality of self-supervised representations.

\vspace{-0.03in}
\subsection{Performance on Downstream Vision Datasets}
\vspace{-0.02in}
Beyond standard vision datasets such as ImageNet, we evaluate SWEET on downstream datasets with limited samples and fine-grained semantics, including Oxford Flowers~\cite{nilsback2008automated} and Stanford Cars~\cite{gebru2017fine}, providing a stringent test of the model’s ability to learn general and discriminative visual representations.

As shown in Table~\ref{tab:downstream}, SWEET achieves superior performance across most settings, suggesting that self-supervised weight templates capture transferable features that generalize well to these challenging tasks.
In contrast, classification-trained templates (e.g., WAVE) and heuristic selection incur larger performance drops, revealing their limited robustness under data scarcity and fine-grained semantic shifts.

\subsection{Ablation and Analysis}
\subsubsection{Effect of Tucker-based Constraints}
\begin{table}[!t]
    \centering
    \setlength{\tabcolsep}{2.7 mm} 
    \caption{Ablation study on Tucker-based constraints.}
    \vspace{-0.1in}
    \resizebox{0.9\linewidth}{!}{
        \begin{tabular}{@{}lc|c|c@{}}
        \toprule[1.5pt]
        & $\textcolor{red!65}{\boldsymbol{L_3}} \textcolor{blue!100}{\boldsymbol{H_{5}}}$ & $\textcolor{red!85}{\boldsymbol{L_4}} \textcolor{blue!85}{\boldsymbol{H_{4}}}$ & $\textcolor{red!100}{\boldsymbol{L_{5}}} \textcolor{blue!65}{\boldsymbol{H_{3}}}$ \\
        \midrule
        w/o Constraits & 57.09 & 57.99 & 54.51 \\
        Linear & 58.09 & 59.00 & 55.66 \\
        Kronecker & 58.83 & 59.47 & 55.88 \\
        \midrule
        \cellcolor{blue!12}{Tucker (Our)} 
               & \cellcolor{blue!12}{\textbf{58.99}} & \cellcolor{blue!12}{\textbf{59.71}} & \cellcolor{blue!12}{\textbf{56.70}} \\
        \bottomrule[1.5pt]
        \end{tabular}   
    }    
    \label{tab:ab_constrait}
\vspace{-0.1in}
\end{table}

Constraint-based pre-training regularizes the optimization process to encapsulate knowledge in a decomposable, scale-agnostic form, enabling modular and scalable knowledge extraction with consistent weight reconstruction across model sizes.
In contrast, as shown in Table~\ref{tab:ab_constrait}, unconstrained pre-training produces tightly coupled, scale-specific weights that generalize poorly to unseen widths or depths.

Compared with linear and Kronecker-based constraints, Tucker-based constraints provide greater flexibility by disentangling knowledge across layer and width dimensions. Linear factorization is limited to a single subspace, while Kronecker decomposition enforces rigid multiplicative structure, both restricting scalability. In contrast, Tucker-based constraints, combined with width-wise stochastic scaling, balance expressiveness and regularization to enable size-agnostic and robust initialization across model scales.

\subsubsection{Effect of Width-wise Stochastic Scaling}
\begin{table}[!t]
    \centering
    \setlength{\tabcolsep}{2.7 mm} 
    \caption{Ablation study on width-wise stochastic scaling (w/o Stoch. Scal.) and architectural enhancements (w/o Arch. Enh.), including SwiGLU, RMSNorm, and RoPE.}
    \vspace{-0.08in}
    \resizebox{0.9 \linewidth}{!}{
        \begin{tabular}{@{}lc|c|c@{}}
        \toprule[1.5pt]
        & $\textcolor{red!65}{\boldsymbol{L_3}} \textcolor{blue!75}{\boldsymbol{H_{3}}}$ & $\textcolor{red!65}{\boldsymbol{L_3}} \textcolor{blue!85}{\boldsymbol{H_{4}}}$ & $\textcolor{red!65}{\boldsymbol{L_3}} \textcolor{blue!100}{\boldsymbol{H_{5}}}$ \\
        \midrule
        w/o Stoch. Scal. & 48.63 & 53.92 & 58.65 \\
        \midrule
        WT-Select & 48.65 & 53.64 & 57.09 \\
        w/o Arch. Enh. & 49.03 & 54.65 & 57.80 \\
        \midrule
        \cellcolor{blue!12}{SWEET} 
               & \cellcolor{blue!12}{\textbf{49.57}} & \cellcolor{blue!12}{\textbf{55.21}} & \cellcolor{blue!12}{\textbf{58.80}} \\
        \bottomrule[1.5pt]
        \end{tabular}   
    }    
    \label{tab:ab_wws}
\vspace{-0.2in}
\end{table}

Width-wise stochastic scaling regularizes knowledge along width-related dimensions by applying structured dropout to weight scalers during pre-training, forcing the model to reorganize information across the width dimension. 
This stochastic perturbation encourages low-index components of templates to capture width-invariant representations.

As shown in Table~\ref{tab:ab_wws}, width-wise stochastic scaling substantially improves the adaptability of learned templates across models of varying widths, consistently outperforming variants without scaling and yielding higher classification accuracy.
We further examine the effect of general architectural enhancements on the structure of weight templates (Sec.~
\ref{sec:wss}). 
Integrating components such as SwiGLU, RMSNorm, and RoPE enhances template expressiveness and stability, promoting consistent feature encoding and improving performance across diverse model configurations.



\vspace{-0.02in}
\subsubsection{Visualization of MAE Reconstruction}
\vspace{-0.02in}
We analyze the learned self-supervised weight templates via MAE-based per-element reconstruction under a 75\% masking ratio, as illustrated in Fig.~\ref{fig:mae}.
Across model scales, the templates consistently reconstruct masked components, indicating that core structural and semantic information is compactly encoded in the unmasked parameters.
Moreover, reconstruction quality remains stable as model size varies, suggesting that the learned templates capture scale-agnostic and reusable representations rather than architecture-specific patterns.
Together, these results provide qualitative evidence that constraint-based pre-training yields coherent and decomposable weight templates, enabling robust reconstruction and transfer across diverse model configurations.

\vspace{-0.02in}
\section{Conclusion}
\vspace{-0.03in}  
We present SWEET, a constraint-based pre-training framework for learning size-agnostic weight templates in a self-supervised manner. 
By combining Tucker-based constraints with width-wise stochastic scaling, SWEET encodes transferable visual knowledge into reusable templates, enabling scalable and task-agnostic model initialization. 
Extensive experiments on \textsc{Classification}, \textsc{Detection}, \textsc{Segmentation}, and \textsc{Generation} show that SWEET consistently outperforms existing scalable initialization methods while maintaining robust performance across diverse model scales.


\section*{Impact Statement}
The broader impact of our work lies in how SWEET redefines vision model initialization through the learning of size-agnostic, self-supervised weight templates that enable efficient initialization across tasks and scales. By offering a unified, scalable, and generalizable initialization paradigm, SWEET has the potential to accelerate research on AI model scaling, improve performance in low-data and specialized settings, and promote more sustainable and flexible industrial AI applications.


\bibliography{icml2026}
\bibliographystyle{icml2026}

\end{document}